# Chaotic Variational Auto Encoder based One Class Classifier for Insurance Fraud Detection

K.S.N.V.K. Gangadhar[1], B. Akhil Kumar[1], Yelleti Vivek[1], Vadlamani Ravi[1*]

[1]Centre for Artificial Intelligence and Machine Learning
Institute for Development and Research in Banking Technology,
Castle Hills Road #1, Masab Tank, Hyderabad-500076, India
ksnvkgangadhar@idrbt.ac.in ; bakhil@idrbt.ac.in; yvivek@idrbt.ac.in; vravi@idrbt.ac.in

**Abstract**

Of late, insurance fraud detection has assumed immense significance owing to the huge financial & reputational losses fraud entails and the phenomenal success of the fraud detection techniques. Insurance is majorly divided into two categories: (i) Life and (ii) Non-life. Non-life insurance in turn includes health insurance and auto insurance among other things. In either of the categories, the fraud detection techniques should be designed in such a way that they capture as many fraudulent transactions as possible. Owing to the rarity of the fraudulent transactions, in this paper, we propose a chaotic variational autoencoder (C-VAE to perform one class classification (OCC) on the genuine transactions. Here, we employed the logistic chaotic map to generate random noise in the latent space. The effectiveness of C-VAE is demonstrated on the health insurance fraud and auto insurance datasets. We considered vanilla Variational Auto Encoder (VAE) as the baseline. It is observed that C-VAE outperformed VAE in both datasets. C-VAE achieved classification rate of 77.9% and 87.25% in health and automobile insurance datasets respectively. Further, t-test conducted at 1% level of significance and 18 degrees of freedom infers that C-VAE is statistically significant than the VAE.

## Declarations:

**Availability of data and material:** All the datasets are publicly available.
**Competing interests:** All the authors declare that they have no competing interests.
**Funding:** Not applicable.

---

[*] Corresponding Author

# 1. Introduction

Insurance policies are designed with the objective of reimbursing losses that occur to individuals due to unfortunate situations. Often, it is observed that a few of the requests from the customers are falsely claimed. This is indeed a fraudulent activity, which entails a huge financial loss to the providing insurance companies/firms. Hence, handling these fraudulent activities not only avoids unnecessary loss but also avoids denting the reputation of the respective companies. Identifying these false insurance claims or fraudulent activity is called Insurance Fraud detection. This is a challenging problem because insurance companies often share very little amount of fraud data. Further, by virtue of its nature, the data is heavily skewed, and is heavily dominated by non-fraudulent samples. Hence, it demands a robust and effective solution mechanism to identify false claims, thereby avoiding the consequences.

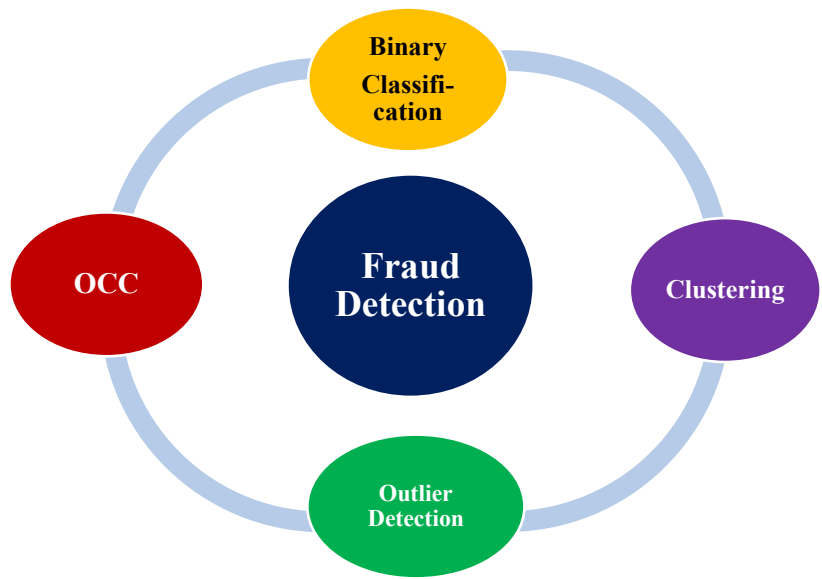

**Fig. 1.** Overview of the Fraud detection methods

In the literature, fraud detection is performed by using various mechanisms (refer to Fig. 1): binary classification, one-class classification (OCC) Clustering and Outlier Detection. In this work, we confined ourselves to the OCC approach. The reasons and distinction between OCC and binary classification is discussed here. In binary classification, machine learning techniques are employed to discriminate into two different classes where positive samples are fraudulent samples. However, in the case of OCC, the machine learning algorithm is trained on the negative sample dataset and tested on the positive sample data. As discussed earlier, the insurance fraud datasets possess very little information pertaining to fraud. Consider an example of providing auto mobile insurance to customers. Here, most of the customers will submit the relevant documents in case of vehicle loss. Then, the respective organization verifies the credibility of the claim and then acts accordingly. Since, most customers claim with a valid reason, but a very few people exist to forge the claim. If such false claims go undetected incur a huge loss to the respective organization. Further, as we can observe that the false claims here are very small in number. Hence, this arises the data imbalance problem. Under such circumstances, binary classification models face huge difficulty in identifying positive samples. Further, sometimes it becomes impractical to employ binary classification models. Considering the above reasons, especially the rarity of the fraudulent dataset, OCC is one of the effective techniques in identifying fraud patterns.

In this study, Insurance Fraud detection is majorly categorized into two different categories: (i) Health care fraud detection, such as medical insurance fraud detection, medical provider fraud detection, and drug fraud detection; (ii) Non-health care fraud detection, such as automobile insurance fraud detection, agent fraud detection etc. There are several studies reported in the literature solving these fraud detection problems by using various binary classification models. But, using OCC is left unexplored, or very few studies reported. Owing to the importance and necessity of OCC models in fraud detection problems, motivated us to work towards it. The fraudulent patterns are so peculiar, and hence discriminating them using a finite training sample is an infeasible solution. Further, the decision boundaries are also significantly nonlinear, especially in the cases of fraud detection datasets. This motivated us to employ Variational Auto Encoder (VAE) as the OCC technique for the effective identification of fraudulent patterns. VAEs are one of the most widely used technique in the OCC in various domains, viz. anomaly detection, classifying deep fakes, outlier detection, novelty detection etc.,

Although VAEs (Kingma and Welling, 2013) accomplish the task, the major hurdle lies in the context of generating some specific datasets. VAEs don't provide control over the generation of the dataset. Further, the disentanglement with the latent dimension also needs to be addressed for effective generation, thereby identifying the fraudulent samples. There are various variants of VAEs proposed by inserting label information in the latent space, implementing more than one latent space, including the priorities of the loss functions etc. However, to the best of our knowledge, there is no study applied to OCC based VAE by modifying the latent space in the context of fraud detection. On the other hand, VAE (Kingma and Welling, 2013) mentioned that different distributions of the latent space could also be used based on the application. The above reasons motivated us to introduce chaotic sequences generated by using chaotic maps, in the latent space. Chaotic maps have the following properties: ergodicity and intrinsically stochastic. These chaotic sequences indeed of different distributions and act as a viable alternative in obtaining control over the generation of datasets. There are various studies where chaotic maps were used successfully in place of the random sequences and proven to be effective in solving the following problems viz., feature subset selection (Vivek et al., 2022), ATM cash demand forecasting (Sarveswara rao et al., 2023), GANs for balancing the data (Kate et al., 2022) etc. This motivated us to incorporate chaotic maps in the VAE and proposed chaotic VAE (C-VAE). Further, as explained earlier, C-VAE is applied in the context of OCC, and named it, Chaotic variational autoencoder (C-VAE).

The major contributions of the current research work are as follows:

- Proposed Chaotic Variational Autoencoder (C-VAE) which incorporates the logistic chaotic map.
- Applied the C-VAE in the context of one class classification.
- C-VAE is applied to solve insurance fraud detection datasets pertaining to both healthcare and non-healthcare domains.
- Developed VAE based OCC and compared its performance with that of C-VAE based OCC

The remainder of the paper is organized out as follows: Section 2 discusses the relevant studies in the literature review. Section 3 discusses the overview of the employed techniques. Section 4 presents proposed Chaotic Variational Auto Encoder. Section 5 describes the dataset. Section 6 discusses the results. Finally, section 7 concludes the paper.

## 2. Literature Review

In this section, we will discuss the relevant studies which are applied to Medicare dataset, and the Automobile insurance dataset. It is observed that majorly, fraud detection is handled in four different mechanisms which are depicted in Fig. 1.

Bauder et al. (2016) proposed a machine learning methodology for healthcare fraud detection problems. The authors employed various machine learning models and performed the analysis using the 5-Fold cross-validation approach. This will help to avoid false claims, and the respective genuine physicians' claims will only be considered. Bauder et al. (2018) proposed an outlier detection methodology and applied it to Medicare dataset. The authors applied various outlier detection algorithms viz., local outlier factor (LOF), autoencoders, isolation forest etc., models. LOF outperformed the rest in detecting the outliers.

Few studies claims that the categorical features presented in the Medicare dataset have a lot of possible values. This indeed results in the generation of high-dimensional datasets. To mitigate this issue, Hancock and Khoshgoftaar (2020), employed CatBoost and LightGBM to perform the encoding. The authors empirically proved that CatBoost yielded a better AUC than the rest of the machine learning (ML) algorithms. Johnson and Khoshgoftaar (2021) focused on the problem of encoding medical provider types. This study mainly aims at proposing procedural-level statistics based encoding techniques to extract useful information. The authors proposed two different methods viz., GloVe and Med-W2V, which uses pre-trained word embeddings.

**Table 1.** Summary of the literature

| Author | Dataset | Employed Technique | Methodology |
|---|---|---|---|
| Hancock and Khoshgoftaar (2020) | Medicare Dataset | CatBoost, LightGBM | Binary Classification |
| Johnson and Khoshgoftaar (2021) | Medicare Dataset | GloVe, Med-W2V | Classification |
| Lavanya and Ganagadaran (2021) | Medicare Dataset | Primitive Sub-peer group analysis (PSPGA) | Clustering |
| Sowah et al. (2019) | Health Insurance claims dataset | Genetic Support Vector Machine (GSVM) | Binary Classification |
| Lavanya and Gangadharan (2021) | Medicare Dataset | Weighted Multi-tree | Binary Classification |
| Bauder and Khosgoftaar (2016) | Temperature Dataset, Medicare Dataset | Probablistic method based on Bayesian interface | Outlier Detection |
| Johnson and Nagrur (2016) | Insurance dataset | Five stage framework | Classification |
| Lavanya et al. (2022) | Medicare Dataset | DADHC | Clustering |
| Braud et al. (2018) | Medicare Dataset | LOF | Outlier Detection |
| Braud et al. (2016) | Medicare Dataset | Multi-nomal Naïve Bayes | Binary Classification |
| Farquad et al. (2012) | Insurance dataset, Credit card churn dataset | SVM + NB / DT | Binary Classification |
| Sundarkumar et al. (2016) | Phishing detection dataset | DT, SVM | Binary Classification |
| Kumar et al. (2015) | insurance fraud detection and credit card churn prediction | K-KNN + SVM | One class classification |
| Nian et al. (2016) | Automobile insurance dataset | SRA | Outlier Detection |
| Liu et al. (2020) | Automobile insurance dataset | Evidential Reasoning + Any Classifier | Binary Classification |
| Itri et al. (2019) | Automobile insurance dataset | Random Forest | Binary Classification |
| Bernand and Vandefful (2014) | Stock market dataset | Optimal prediction intervals | Predictive Intervals |
| Phua et al. (2004) | Automobile insurance dataset | Stacking + Bagging | Binary Classification |
| **Current Study** | **Automobile insurance dataset and Medicare dataset** | **C-VAE** | **One Class Classification** |

Behavioural patterns possess critical information pertaining to the patient's help in identifying fraud detection. Lavanya and Gangadharan (2021) developed a primitive sub-peer group analysis (PSPGA) which flags suspicious behaviour, thereby avoiding false claims. PSPGA is inspired by an unsupervised learning algorithm, peer group analysis. PSPGA adapts itself by considering the updates from the local peer groups. This makes PSPGA highly adaptive to the underlying drifts. Lavanya et al. (2022) further proposed a novel approach to analyze the hidden patterns which contain critical information related to fraudulent activities. The authors named it to drift analysis on decomposed health care claims (DADHC), where at the first stage pseudo additive decomposition method is used to decompose the healthcare claims. Later, topological clustering, based on adaptive resonance theory, is used to analyze the patterns.

Sowah et al. (2019) proposed a genetic support vector machine (GSVM), which is a novel technique which is dependent on genetic support vector machines. GSVM identifies the anomalies and is used to identify fraud. The authors validated their approach by applying it to the publicly available dataset National Healthcare insurance claims dataset. When combined with SVM kernels, GSVM attained better performance.

Automated Labelling and provider profiling are among the most ignored yet important problems to be addressed, especially in fraud detection. Lavanya and Gangadharan (2021) addressed the above two problems by using the weighted multi-tree. This approach constructs a directed acyclic graph (DAG) which avoids ambiguity.

Bauder and Khosgoftaar (2016) proposed a probabilistic-based outlier detection algorithm to identify fraud in the fraud detection problems. This algorithm is based on a Bayesian interface. Along with the probability scores, it also provides credibility scores, thereby increasing the confidence of being a fraud. The authors applied to two different case studies, (i) temperature data and (ii) the medicare dataset. The authors claimed that their proposed approach effectively identified outliers.

Johnson and Nagrur (2016) proposed a five-stage methodology to mitigate fraud effectively. The first stages primarily focus on the providers, services and claim amounts. Later, in stage 4, the overall risk measure is calculated by using the extracted information from the first three stages. At the end of stage 5, the decision tree is employed to classify the fraudulent activities by considering the information from the first four stages and the threshold.

Fraquad et al. (2012) proposed a modified active learning framework where the SVM is used to extract the if-then rules and applied to solve churn prediction and insurance fraud detection. The authors proposed a three-phase approach which is described as follows: (i) SVM is used for the recursive feature elimination, (ii) synthetic data is generated by using active learning, and (iii) by using DT and NB, the rules are generated which are used to identify whether the fraud has happened or not.

Detecting malware constitutes several challenges and demands a strong, sophisticated defence mechanism. In general, malware attackers use application program interface calls (API) to steal credit card numbers, critical personal information, etc. Hence, Sundarkumar et al. (2015) used this API call information and applied topic modelling techniques to extract the information. This is followed by various ML models for malware detection. Their results prove that DT and SVM outperformed the rest, while DT is the winner in the interpretability aspect.

Data imbalance in fraud detection is one of the key challenges to be addressed effectively. Hence, Kumar and Ravi (2015) proposed a novel hybrid approach which employs K-Reverse Nearest Neighbourhood and One class Support vector machine (OCSVM). The authors validated the proposed approach on Auto mobile insurance fraud detection and credit card churn prediction datasets. The balancing approach achieved better results than the imbalanced.

Nian et al. (2016) proposed an unsupervised learning based algorithm to rank the anomalies thereby identifying whether the fraud happened or not. For this, the authors used the spectral ranking method for the anomaly (SRA). SRA generates the non-principal eigenvectors which are used to rank the anomalies. The authors demonstrated that the first non-principal eigenvector has the ability to bifurcate the anomalies and normal.

Liu et al. (2020) proposed a data-driven inferential modelling technique which combines evidential reasoning and the probability of being a fraud. Evidential reasoning composes a set of evidence and then each one is weighted, and then the decision is taken conjunctively. The authors applied their approach to the automobile insurance dataset, which outperformed logistic regression and random forest. Itri et al. (2019) conducted an extensive study which focuses on the effectiveness and verifiability of machine learning algorithms. The authors demonstrated that random forest achieved the best performance. An optimal strategy is always helpful in maintaining efficient portfolios. Aiming this, Bernard and Vandefful (2014) obtained the mean and variance of the optimal strategy, which will help to predict fraud detection.

Phua et al. (2004) proposed a hybrid technique comprising stacking and bagging to efficiently handle the skewed distributions, which is more relevant in fraud detection datasets. The authors claimed that their proposed hybrid approach outperformed the C4.5 and random forest.

## 3. Overview of the techniques employed

### 3.1 Variational Auto Encoder

Variational Auto Encoder (VAE), proposed by Knigma and Welling (2013), is an unsupervised learning method. It provides a probabilistic manner for describing an observation in latent space. The major objective is to minimize the difference between a supposed distribution and the original distribution. It can be used in various applications such as dimensionality reduction, image denoising, one-class classification, outlier detection, fraud detection etc.

The architecture of VAE mainly comprises an encoder and decoder. The encoder aims to learn the data encoding efficiently and passes it to the bottleneck architecture. Then, the decoder utilizes the latent space in the bottleneck layer and regenerates the dataset. Then, the backpropagation is applied in the form of the loss function. The loss function comprises the loss generated between the original dataset and the generated dataset and the approximate distribution of the generated dataset. In general, VAE uses MSE and KL-divergence as the loss function to measure the closeness of the dataset.

**3.2 Chaos Theory & Chaotic Maps**

Lorenz (1963) proposed the chaos theory, which is a dynamic deterministic system. He described chaos as follows: "*When the present determines the future, but the approximate present does not approximately determine the future*". It evolves from initial conditions, and the state space is defined by using trajectories generated in the system. There are two vital properties which make chaotic maps more adaptive are as follows: ergodicity and intrinsically stochastic in nature.

Chaotic maps are a combination of sequences of numbers and are designed in such a way that they exhibit the above chaotic properties. Researchers have used it as an alternative to the pseudo-random numbers in various useful applications such as feature subset selection (Vivek et al., 2022), handling data imbalance problems, and subsumed with EAs to improve adaptability (Vivek et al., 2022) etc. For instance, the pseudo-random number generated by the logistic map turns out to be following the exponentiated Weibull distribution. We performed Kolgomorgov Smiron (KS) test[1] to know the underlying logistic map distribution. In some sense, chaos represents deterministic randomness. It is deterministic because we can predict the sequence of numbers being generated as they are governed by differential equations and hence change subject to the initial conditions. In the current study, we used a logistic chaotic map which is described as follows:

**Logistic Map** It's a discrete-time demographic model and exhibits chaotic behaviour (refer to Eq. 1). The following map is a polynomial mapping and has degree 2.

$$x_{t+1} = \lambda * (x_t * (1 - x_t)) \tag{1}$$

---

[1] https://docs.scipy.org/doc/scipy/index.html

Here, the constant $\lambda$ lies in the range of [0, 4] and determines the behaviour of this Logistic map. If $3.56 < \lambda < 4$, it is observed to have chaotic behaviour. In the current research study, $\lambda = 4$ is chosen.

## 4. Proposed C-VAE

In this section, we will first discuss the architecture of the C-VAE, which is followed by the discussion of training and testing of the proposed C-VAE.

### 4.1 Architecture of C-VAE

The architecture of the proposed chaotic variational autoencoder (C-VAE) is depicted in Fig. 2. Likewise VAE, it also has three different components, viz., Encoder, Latent distribution and Decoder, respectively. C-VAE is different from VAE in the latent distribution component. Each of the components are described as follows:

**Encoder**

The encoder of the C-VAE is represented by $\phi$. It receives the complete dataset of size $n*nf$, where n is the number of samples and $nf$ is the number of features and generates the lower dimensional vector. This lower dimensional vector is called, the latent vector of size $n*z$, where $z$ is the dimension of the latent space. The encoder learns the mean and variance of the distribution of $z$. In the proposed C-VAE, we employed a multi layer perceptron (MLP) as the encoder.

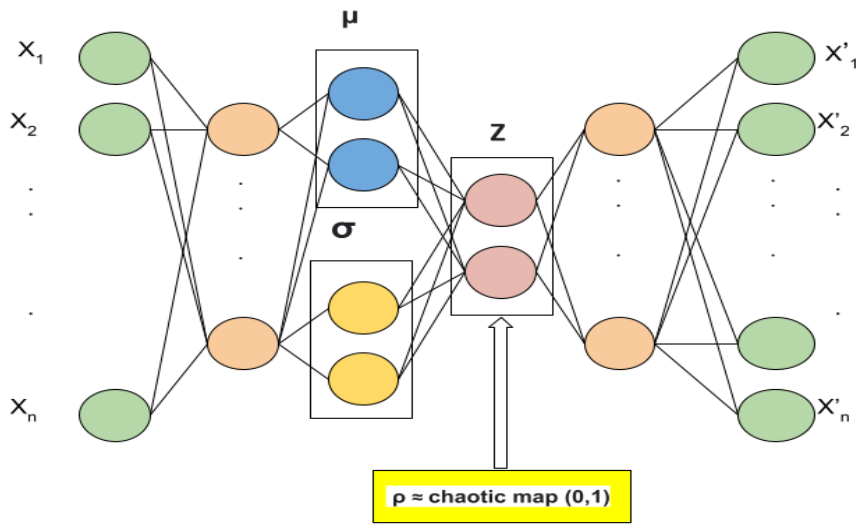

**Fig. 2.** Architecture of Chaotic Variational Auto Encoder

**Latent distribution**

The random noise which is present in the latent distribution, contains the chaotic values generated by using the underlying chaotic map. In our study, we used a logistic map. The initial seed for the logistic map is a random number. Thereafter, the numbers in the chaotic series are generated by using logistic map (refer to Eq. 1). As explained earlier, we observed that the logistic map turns out to be following the exponentiated-Weibull distribution. The distribution is found by performing Kolgomorgov Smiron (KS) test.

**Decoder**

The decoder of the C-VAE is represented by $\theta$. It receives the sample latent vectors of size $n*z$, where n is the number of samples and $z$ is the dimensions of latent space. This lower dimensional vector is utilized to reproduce the dataset thereby minimizing the loss. This loss function comprises two important optimization tasks: (i) reconstruction error: which calculates the difference in the original and generated dataset. (ii) minimize KL divergence: distribution of latent space vectors to the normality.

Here, in C-VAE, we employed a multi-layer perceptron (MLP) as the decoder.

## 4.2 C-VAE

In this section, we will discuss the training and test procedure that is adopted in the current study. The sequence of steps to be adopted is presented in Algorithm 1 and is depicted in Fig. 3. The model training is performed from steps 1-11, and the model test starts from step 12 onwards.

The algorithm starts by dividing the training and test datasets. In one class classification, the training has to be performed only on negative samples dataset, and the test has to be done on only positive class. The same is performed in the first two steps of Algorithm 1. Also, the dataset has to be normalized. This is followed by initializing the C-VAE parameters. The model is represented by using $M_{[\phi,\theta]}$, where $\phi$ is the encoder and $\theta$ is the decoder of the model. Then the model $M_{[\phi,\theta]}$ has to be trained for the specified $L$ number of epochs.

$$loss = MSE(X, X') + KL\_divergence(q_\theta(Z), p(Z)) \quad (2)$$

Where, $X$ is the original dataset, $X'$ is the generated dataset, $Z$ is the latent vector, $q_\theta(Z)$ is a variational posterior probability, and $p(Z)$ is a prior probability.

Depending on batchsize, randomly a subset of the dataset is obtained. Henceforth, the encoder is trained thereby generating the latent space $Z$ vector. As discussed earlier in the architecture, the encoder learns the mean and variance and then generates the $Z$ vector. Thereafter, the noise is generated by chaotic maps, $\rho$ is used to reconstruct the latent vector and then given as the input for the decoder.

**Algorithm 1.** Pseudocode of the proposed C-VAE

*Input:* L: number of epochs, lr: Learning Rate, mom: Momentum
*Output:* $M_{[\phi,\theta]}$: trained model, test_cr scores: Test classification rate scores

1. $X_{train} \leftarrow$ Normalize ( $X_{[class=0]}$)
2. $X_{test} \leftarrow$ Normalize ( $X_{[class=1]}$)
3. $M_{[\phi,\theta]} \leftarrow \emptyset$     // initialize the C-VAE model;
4. **for** k is 1 **to** L **do**
5.     x ← Randomly obtain the dataset from $X_{train}$
6.     ϵ ← Train the Encoder model
7.     ρ ← Generate random noise by using Logistic Map
8.     ζ ← Train the Decoder model by using chaotic random noise and Z
9.     Calculate Loss (refer to Eq. 2)
10.    Perform Backpropagation
         // update model $M_{[\phi,\theta]}$ parameters
11. **end for**
12. $X'_{test} \leftarrow$ Test the trained model, $M_{[\phi,\theta]}$ performance on $X_{test}$
13. $Decision_{scores} \leftarrow$ Compute difference between $X_{test}$ and $X'_{test}$
14. **for** score **in** $Decision_{scores}$
15.    **if** score > threshold:
             Classify it as Positive sample
16.    **else**
17.        Classify it as Negative sample
18. Calculate CR score (refer to Eq. 3)
19. **return** $M_{[\phi,\theta]}$, test_cr scores

Now, the decoder learns the underlying pattern and then regenerates the vector, in such a way that it is minimizing the loss function. The loss functions majorly contain two components (refer to Section 4.1). In our case, we used MSE and KL-Divergence as the loss function components. Now, the loss is computed accordingly, and then backpropagation is invoked. This will update the $M_{[\phi,\theta]}$ parameters. The above process is repeated for the specified *L* number of epochs.

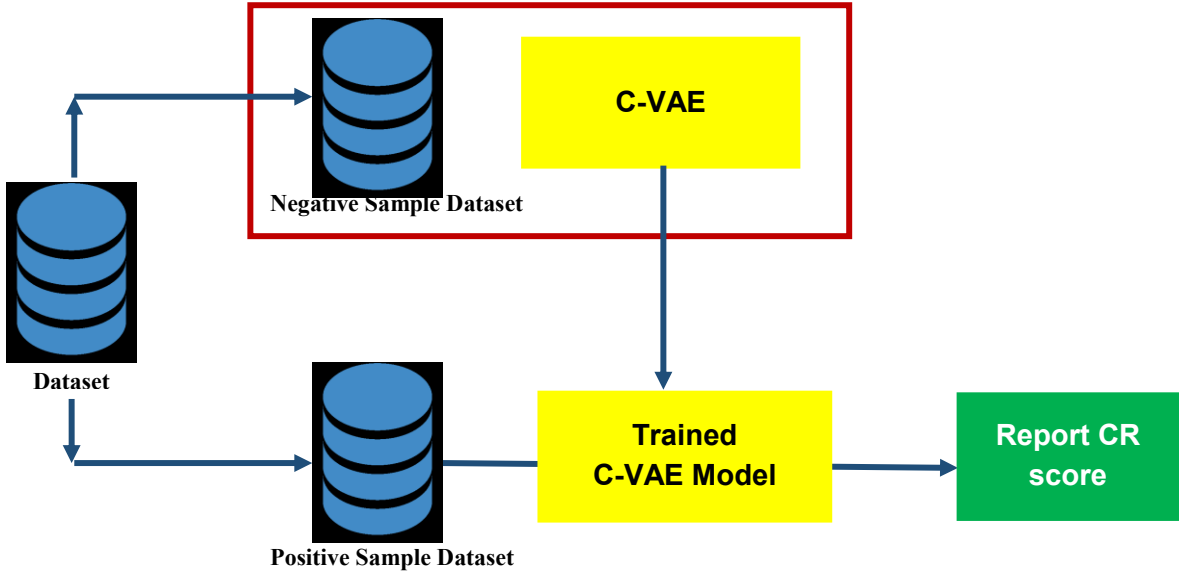

**Fig. 3.** Schematic diagram for the One Class Classification

Once the model, $M_{[\phi,\theta]}$ is trained, then the performance of the model is evaluated on the test dataset, $X_{test}$. This result in the generation of $X'_{test}$, which are regenerated vectors of the test dataset. Now, the decision scores are calculated by computing the difference between the generated vector and the original vector. It is evident that the fraudulent samples are the tampered ones. Hence, C-VAE will find it difficult to regenerate them. The same is also observed for VAE. Hence, the fraudulent (or positive) class samples show more dissimilarity and have more decision scores than the non-fraudulent (or negative) class samples. The value of the decision scores is always positive. But '*how far is too far*' is still a question. This is where the threshold is used to decide whether the data point belongs to either a fraudulent or non-fraudulent class. It is to be noted that, the threshold is user-defined. In our study, we accommodated the threshold as *n\*0.01*, where n is the number of samples. Now, steps 14-18 are invoked where the threshold decides the type of class a sample belongs to. This is followed by calculating the test CR score. Now, the trained model and the corresponding CR score are returned.

## 5 Dataset Description

As discussed earlier, we performed both healthcare and non-healthcare categories. Hence, we make use of the Medicare dataset from the healthcare domain and Auto insurance fraud detection from the non-healthcare insurance category. It is observed that both these datasets are imbalanced and hence applicable to applying techniques. The description of the details was discussed as follows:

## 5.1 Medicare Dataset

This is a publicly available dataset, collected from the Center for Medicare Services (CMS)[2,3] during the period 2017-2019. It comprises the medicare provider utilization and the payment data for physicians and other suppliers. It also includes the annual claims. This dataset has both categorical and numerical data types. The features, label and corresponding description are presented in Table 2. The data availed the health common procedure coding system, which identifies each provider by a healthcare procedure code (HCPS). These codes are availed by CMS to guarantee the systematic and uniform processing of insurance claims and payments.

The dataset preprocessing is performed as follows: dataset provided by CMS is unlabelled. But CMS also maintains the LEIE database, which contains information about doctors and other health care providers prohibited from practice. As discussed earlier, each of the provider and doctor is recognized by a unique identical number. If a provider or doctor prohibited from practice indicate that they might involved in some fraudulent category. Hence, we utilized this information and added 'exclusion' as the class label which indicates whether the fraud happened or not. Whomever the doctors and providers are prohibited from practice are considered a fraud class (positive class) and others as a non-fraud class (negative class). By utilizing the LEIE database[4], we converted this unlabelled dataset into a labelled dataset. This avails one to apply the supervised techniques.

**Table 2.** Description of the features and class label of the Medicare Part B dataset

| Name of the Feature and Class Label | Description | Type of the feature |
|---|---|---|
| *npi* | Unique provider identification number | Categorical |
| *provider_type* | Medical provider's specialization | Categorical |
| *nppes_provider_gender* | Gender of the medical provider | Categorical |
| *line_srvc_cnt* | Number of providers/services performed by the medical provider | Numerical |
| *bene_unique_cnt* | Number of distinct medicare beneficiaries receiving the service | Numerical |
| *bene_day_srvc_cnt* | Average of the charges that the provider submitted for the service | Numerical |
| *average_submitted_chrg_amnt* | Average payment claimed by the provider for the service | Numerical |
| *average_medicare_payment_amt* | Average payment performed by the provider against the claim o | Numerical |
| *exclusion* | Fraud labels obtained from LEIE database Categorical | Class Label |

---

[2] https://www.Medicare.gov
[3] https://www.cms.gov/Research-Statistics-Data-and-Systems/Statistics-Trends-andReports/NationalHealthExpendData/Downloads/highlights.pdf
[4] https://oig.hhs.gov/exclusions/index.asp

All the categorical features are handled by using one hot encoding technique and the numerical features are normalized. After incorporating the '*exclusion*' class label, the dataset turned into a binary classification dataset. The dataset contains 8304 cases, out of which 895 are found to be fraudulent, and 7409 are non-fraudulent. It contains 8 features out of which 3 are categorical features, and 5 are numerical features. This dataset is imbalanced in nature.

Table 3. Description of the features and class label of Auto mobile Insurance dataset

| Name of the Features and Class Label | Description | Type of the feature |
|---|---|---|
| *Month* | Month in which insurance is issued | Categorical |
| *Week of Month* | Week of the corresponding month in which insurance is issued | Numerical |
| *Day of week* | Day in which insurance is issued | Categorical |
| *Month Claimed* | Month in which insurance is claimed | Categorical |
| *Week of Month Claimed* | Week of the corresponding month in which insurance is claimed | Numerical |
| *Day of week Claimed* | Day in which insurance is claimed | Categorical |
| *Year* | Year of issue | Categorical |
| *Make* | Name of the vehicle | Categorical |
| *Accident Area* | Whether accident occurred in rural or urban | Categorical |
| *Sex* | Gender of the customer | Categorical |
| *Martial Status* | Current martial status of the customer | Categorical |
| *Age* | Age of the customer | Categorical |
| *Fault* | Type of fault (policyholder/ third party) | Categorical |
| *Policy Type* | Type of the insurance policy | Categorical |
| *Vehicle Category* | Category of the vehicle | Categorical |
| *Vehicle Price* | Price of the vehicle (cost price) | Categorical |
| *Policy Number* | ID of the policy | Categorical |
| *Rep Number* | ID of the insurance representative | Categorical |
| *Deductible* | Whether the money is deductible or not | Numerical |
| *Driver Rating* | Rating possessed by the driver | Numerical |
| *Days policy claimed* | Days after which policy is claimed | Categorical |
| *Days policy accident* | Days before accident is happened | Categorical |
| *Past Number of Claims* | Represents the number of past claims | Numerical |
| *Age of Vehicle* | Tenure of the vehicle till date | Categorical |
| *Age of Policy Holder* | Age of the insurance policy holder | Categorical |
| *Police Report Filed* | Whether the police report filed against the accident | Categorical |
| *Witness Present* | Anyone witnessed the accident | Categorical |
| *Agent Type* | Whether the agent is an internal / external | Categorical |
| *Number of Supplements* | Total number of supplements | Categorical |
| *Address change claim* | Months after which the address is changed | Categorical |
| *Number of cars* | Number of vehicles one possess | Numerical |
| *Base policy* | Type of the insurance policy | Categorical |
| *Fraud Found* | Fraud is detected or not | Class Label |

## 5.2 Auto Mobile Insurance dataset

This is the only fraud detection dataset related to the automobile insurance domain. This dataset is provided by Angoss knowledge seeker software. The following dataset is obtained from Pyle, and the complete description of the features and class label is presented in Table 3.

The dataset contains the insurance claim information against the accidents that happened during the period of 1994 to 1996. It contains 11,338 records in total and comprises 25 categorical features and 6 numerical features. All these categorical features are handled by using one-hot encoding, and ordinal encoding wherever it is applicable. This dataset is also imbalanced, where 6% of 11,338 records are fraudulent claims, and the rest, 94% are legitimate records.

## 5.3 Evaluation measure

While performing OCC, either during the training or test phase, only one class is provided. Hence, here we accommodated the classification (CR) metric to assess how many fraudulent samples are identified by the underlying OCC model.

**Classification Rate**

It is the ratio of the number of points correctly identified to be the positive class sample by an OCC model out of the total positive samples. It is mathematically represented as given in Eq. (3).

$$Classification\ rate\ (CR) = \frac{Number\ of\ correctly\ classified\ positive\ samples * 100}{Total\ number\ of\ positive\ samples} \quad (3)$$

Table 4. Hyperparameters for all the techniques

| Model | Hyperparameters |
|---|---|
| VAE | **Both Encoder and Decoder**<br>Learning rate: [0.001,0.0005]<br>Momentum: [0.005,0.007,0.009]<br>Epochs: [50,75,100,250,150]<br>Activation: ['relu', 'tanh', 'leakyRelu']<br>Optimizers: ['Adam', 'SGD']<br>#Number of layers: [5,7] |
| C-VAE | **Both Encoder and Decoder**<br>Learning rate: [0.001,0.0005]<br>Momentum: [0.005,0.007,0.009]<br>Epochs: [50,75,100,250,150]<br>Activation: ['relu', 'tanh', 'leakyRelu']<br>Optimizers: ['Adam', 'SGD']<br>#Number of layers: [5, 7] |

# 6 Results and discussion

We observed that the random seed influences the performance of VAE, especially after introducing the chaotic maps. Hence, to remove the effect of the randomness, we performed the following experiment for 15 runs using the Hyper-parameters (refer to Table 4). The best CR scores obtained in each run are considered and then calculated the mean and standard deviation. The mean CR score and standard deviation achieved by each model are presented in Table 5. Mean represents the average performance achieved by each of the model in different runs and the standard deviation refers to the variability in the model performance.

It is observed that, C-VAE outperformed VAE in both Medicare and Automobile insurance datasets in terms of mean CR scores. The higher mean CR scores is always desired. C-VAE attained a CR score of 77.9 in the medicare dataset and an 87.25 CR score in the automobile insurance dataset. However, VAE attained 73.13 and 86.9, CR scores respectively.

**Table 5.** Classification Rate obtained by VAE and C-VAE

| Dataset | Model | Mean Classification Rate (Standard Deviation) |
|---|---|---|
| Medicare | VAE | 73.13 (0.05) |
|  | C-VAE | **77.9 ( 0.36)** |
| Automobile insurance | VAE | 86.9 (0.02) |
|  | C-VAE | **87.25 ( 0.08)** |

As mentioned earlier, the more the standard deviation, the more the model performance is varying. Hence, the lesser standard deviation is always desired. It is observed that C-VAE has more standard deviation than the VAE. This says that VAE is more stable than C-VAE. However, the difference is very marginal and it is observed that the lower limit of C-VAE throughout the 15 runs is more than the maximum CR score attained by VAE. Hence, overall, C-VAE outperformed VAE. The results presented in Table 5, indicates that after introducing the chaotic maps, C-VAE became more powerful in discriminating the positive and negative samples. Hence, this is indeed superior to the VAE. We observed that the logistic map follows Weibull distribution (refer to section 3.2). However, the random numbers follow the normal distribution. Hence, it is derived that employing chaotic maps is more powerful than the normal distribution.

## 6.1 Statistical testing of the results

Further, we conducted the t-test analysis to statistically prove whether the superiority of the performance of the models is purely a coincidence or due to its superior nature.

The following t-test analysis is conducted on CR scores obtained by the respective OCC models throughout the 15 runs. The degrees of freedom is 28 (=15+15-2). From Table 6, we observed that in both the datasets, the p-value is smaller than the level of significance (5%). Hence, the null hypothesis is accepted. Hence, C-VAE and VAE are statistically different to each other. C-VAE is shown to be significantly better than VAE.

Table 6. T-test analysis conducted between VAE and C-VAE

| Dataset | Model | t-statistic | p-value |
|---|---|---|---|
| Medicare | VAE vs C-VAE | 48.25 | $1.67 \times 10^{-28}$ |
| Automobile insurance | VAE vs C-VAE | 14.11 | $2.95 \times 10^{-14}$ |

## 7. Conclusions and Limitations

This is a first-of-its-kind study, where we proposed a chaotic VAE (C-VAE), by employing the logistic chaotic map in the latent space. The significance of C-VAE is proved by applying it to the Insurance Fraud detection domain i.e., the Medicare dataset from the healthcare category and Auto insurance from the non-healthcare category. It is observed that C-VAE outperformed VAE in both datasets. C-VAE achieved CR scores of 77.9 and 87.25 in Medicare and Automobile Insurance datasets respectively. The t-test analysis is further conducted which proves that C-VAE is more statistically significant than the VAE.

Now, we will discuss a few important future directions. In the study, we only explored one chaotic map, the other chaotic maps can also be further explored. We designed C-VAE by considering VAE. However, the other variants of VAE can also be incorporated with these chaotic maps is also a potential future direction. The current study is applied to one class classification problem. It can be further extended to other applications such as dimensionality reduction, image denoising, privacy-preserving data mining, etc.